\documentclass[11pt,a4paper,copyright]{google}

\usepackage[authoryear,sort&compress,round]{natbib}
\usepackage{hyperref}
\usepackage{url}
\usepackage{subcaption}
\usepackage{nicefrac}
\usepackage{pifont}
\usepackage{xspace}
\usepackage{listings}
\lstdefinestyle{cfg}{
  basicstyle=\ttfamily\footnotesize,
  commentstyle=\color{gray},
  breaklines=true,
  frame=single,
  framesep=5pt,
  columns=fullflexible,
  keepspaces=true,
  showstringspaces=false,
}
\graphicspath{{figures/}{assets/}}

\hypersetup{
  breaklinks=true,
  colorlinks=true,
  linkcolor=violet,
  urlcolor=blue,
  citecolor=purple
}

\title{AQuA: Recursively Self-Improving Quantitative Trading
Research Agents}

\author{%
\begin{minipage}{\linewidth}
\centering
{\bfseries
Jiacheng Guo\textsuperscript{1*},
Suozhi Huang\textsuperscript{1*},
Yunlong Gao\textsuperscript{2*},
Zihao Li\textsuperscript{1},\\
Jason Ge,
Shushu Liang\textsuperscript{3},
Xu Kuang\textsuperscript{4},
Mengdi Wang\textsuperscript{1}\\[4pt]
}
{\small\normalfont
\textsuperscript{1}Princeton University \quad
\textsuperscript{2}Ant Group\\[2pt]
\textsuperscript{3}Harvard University \quad
\textsuperscript{4}Stanford University
}
\end{minipage}%
}

\renewcommand{\firstpageleftfooter}{\textsuperscript{*}Equal contribution}

\begin{document}

\begin{abstract}
We study recursive self-improvement at the level of quantitative-investment research: whether an
autonomous system can use evidence from earlier experiments to improve the hypotheses and candidates
proposed in later iterations. We present \textbf{AQuA}, which comprises two separate
language-model-driven research systems: one for symbolic factor discovery and one for trainable
model development. Each system records experimental results and uses them to guide subsequent proposals.
Each operates in a fixed sandbox,
which fixes the data splits, feature and label definitions, and evaluator while allowing the model to
act only through constrained factor expressions or configuration diffs. The factor system, a
manager-mediated multi-agent pipeline, discovers and combines factors into a signal that reaches a
combined validation information coefficient of about $0.190$ on a crypto universe. The model system, a
config-driven loop over a hybrid time-series architecture, reaches a per-stock information
coefficient of $+0.0843$ on US equities and converts it into a threshold long/short strategy with a
held-out Sharpe of up to $+2.50$ at a two-leg cost. The strategy is positive in every year from 2021
to 2025.
\end{abstract}

\maketitle

\section{Introduction}
\label{sec:intro}


Quantitative-investment research searches over a large space of factors and models, and small
methodological errors can turn into convincing but non-reproducible backtests. A feature that reads
future information, a strategy selected on the test set, or a result confined to one favorable
regime may all fail out of sample~\cite{bailey2014pseudo,bailey2017probability}. Quantitative
research therefore relies on frozen data splits, held-out evaluation, and skepticism toward results
that look unusually strong~\cite{harvey2016cross}.
The underlying observations also require scrutiny: research on local officials and GDP-data
manipulation examines night lights as an alternative source of economic
information~\cite{liang2017gdp}. Data provenance and measurement therefore matter alongside
experimental design when assessing quantitative evidence.

Large language models can now propose hypotheses, write experiments, and revise their search from
empirical feedback. Existing quantitative agents, however, generally focus on either factor
discovery~\cite{shi2026alphajungle,chen2025multiagentalpha} or model
development~\cite{kabir2025lstmtransformer,song2025transformerrl}. More importantly, an unconstrained
agent can corrupt the evidence on which its later iterations depend.

A code-generating agent may inadvertently introduce a temporal-alignment or preprocessing error that
uses information unavailable at prediction time. A reviewing agent may miss the bug because the code
appears semantically plausible. If the resulting leakage produces a high score, the experiment may
be stored as a successful precedent and propagated through later iterations. Recursive improvement
can therefore amplify an undetected error as readily as a genuine discovery.

Prompt-level instructions and model-based review do not provide a reliable integrity boundary:
repeated access to a fixed holdout can cause adaptive overfitting~\cite{blum2015ladder}, and
language-model agents have been observed exploiting misspecified objectives, tests, evaluators, or
reward mechanisms~\cite{denison2024sycophancy,baker2025monitoring,atinafu2026rewardhacking}.

AQuA restricts the experimental changes available to the agent. Each system fixes its data
pipeline, splits, labels, and evaluator before autonomous iteration begins. The agent cannot write
arbitrary experimental code; it can only emit a program in a restricted domain-specific language
(DSL), which contains no operation for modifying or bypassing the sealed data path or evaluator.

We build \textbf{AQuA}, comprising two separate recursively self-improving research systems: one for
factor discovery and one for model development. In each system, experimental results update a research record that guides later proposals.
The data pipeline and evaluation protocol remain fixed throughout the search.

\begin{figure}[t]
  \centering
  \includegraphics[width=\textwidth]{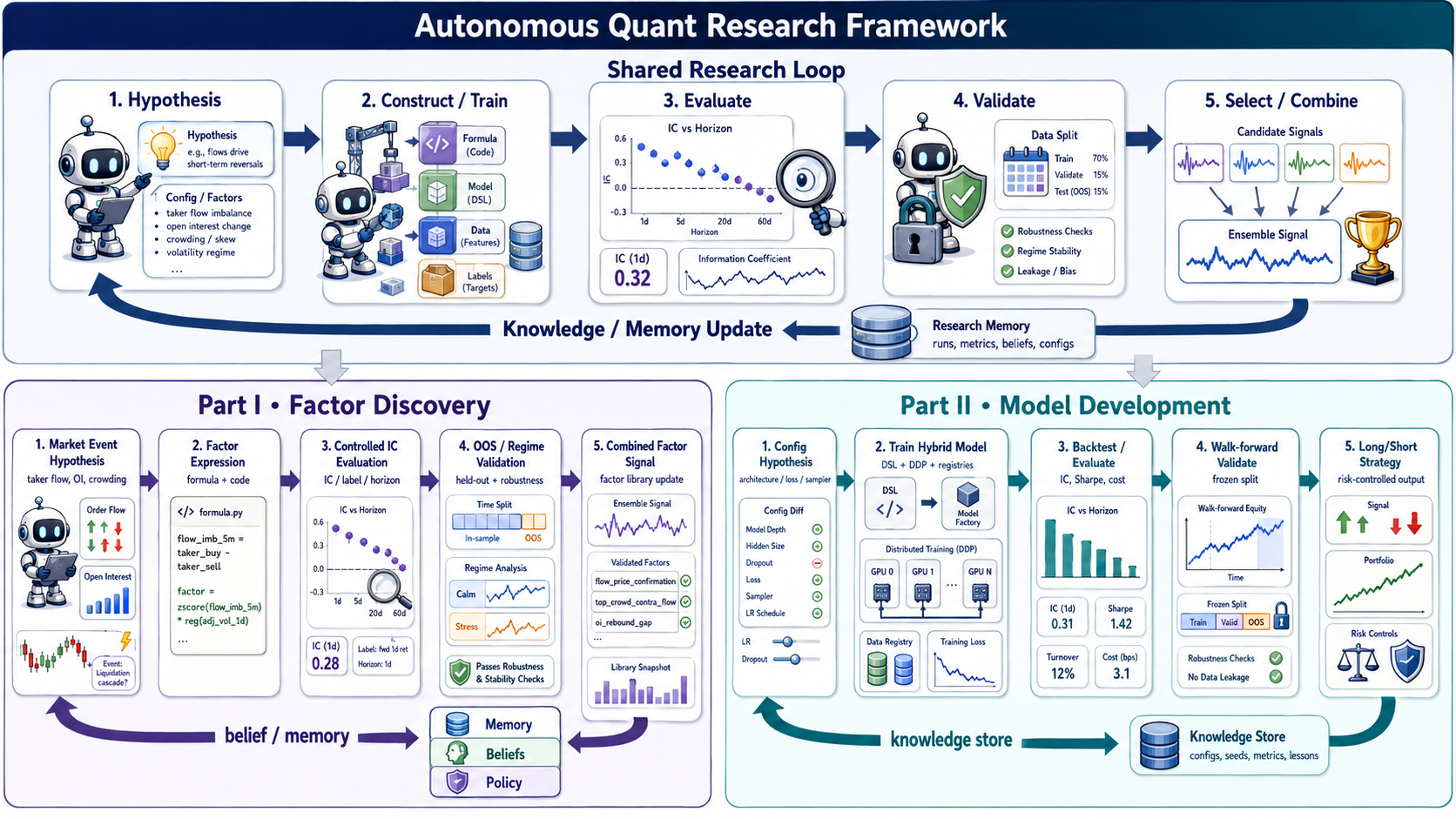}
  \caption{Overview of the two AQuA systems. Factor discovery (Part~I) and model development
  (Part~II) use separate agents, memories, search spaces, and outputs. Each closes its own loop
  through hypothesis, construct/train, evaluate, validate, select/combine, and a persistent update
  that guides the next iteration.}
  \label{fig:overview}
\end{figure}


AQuA searches over factor expressions and model configurations through a restricted DSL.
The agent can select registered operations and their parameters; the harness controls data access
and evaluation. Part~I implements this principle with a manager-mediated
multi-agent pipeline that proposes falsifiable economic hypotheses, evaluates factors, combines
surviving signals, and carries beliefs across runs. Part~II implements it with a config-driven loop
over a hybrid time-series model; each configuration diff defines one comparable model variant, and
its outcome updates the knowledge used to propose the next variant.

The two systems yield improvements in their respective evaluations. On a crypto universe,
Part~I reaches a combined validation information coefficient of about $0.190$. On US equities, Part~II reaches a per-stock information
coefficient of $+0.0843$, versus $+0.0613$ for the strongest baseline, a GRU, under the same
evaluator---an absolute improvement of $+0.0230$ and a relative improvement of $37.5\%$. The
resulting threshold long/short strategy reaches a held-out Sharpe of $+2.50$ at a two-leg cost of
$2$\,bps and is positive in every year from 2021 to 2025. A stricter walk-forward evaluation, in
which every model and strategy parameter is fixed using only data available before the next test
segment, retains a Sharpe of about $+2.0$.

Our main contributions are:
\begin{itemize}
  \item \textbf{We instantiate recursive self-improvement in two separate components of
  quantitative research.} In both factor discovery and model development, records of evaluated candidates
  guide subsequent proposals within each system.
  \item \textbf{We build an autonomous factor-discovery system.} A manager-mediated multi-agent
  pipeline proposes, evaluates, and combines factors while carrying empirical beliefs across runs.
  \item \textbf{We design a hybrid time-series model and an autonomous development loop.} A
  config-driven agent trains directly comparable variants under a sealed evaluation sandbox.
  \item \textbf{We evaluate the resulting factors and models.} We report combined-factor validation
  IC in crypto and held-out predictive and trading performance in US equities. The equity strategy remains profitable
  under a fully causal walk-forward evaluation.
\end{itemize}

Section~\ref{sec:method} describes the shared high-level pattern and the two sealed sandboxes;
Sections~\ref{sec:part1} and~\ref{sec:part2} present factor discovery and model development.

\section{Related Work}
\label{sec:related}

\paragraph{Economic hypotheses and asset pricing.}
Economic evidence provides hypotheses for quantitative factor research.
\citet{chen2022chasing} connect retail investors' return-chasing propensities to subsequent
investor and stock returns, while \citet{chen2023bubble} study how fundamentals, investor entry,
and shifts in preferences or beliefs contribute to returns during a stock-market bubble.
\citet{liang2023essays} examines heterogeneous retail-investor behavior and its price effects.
In catastrophe-bond markets, \citet{liang2025catastrophe} relate prices to expected losses,
issuer characteristics, and credit-market conditions. These studies illustrate how behavioral
and institutional mechanisms can motivate testable hypotheses, the starting point of our
factor-discovery loop.

\paragraph{LLM-driven alpha mining.}
A fast-growing line of work uses large language models to mine formulaic alpha factors. It builds on
operator-based factor search by genetic programming~\cite{zhang2020autoalpha, cui2021alphaevolve} and
reinforcement learning~\cite{yu2023synergistic, zhang2026feedback, zhao2025quantfactor, zhao2025trajectory}, and now spans evolutionary and
agentic search~\cite{han2026quantaalpha, alphaagentevo2026, tang2025alphaagent, wu2026evoalpha,
huang2026hypotheses, yi2026alphaschema, liu2025cogalpha, yu2026autonomousalpha}, program-level
synthesis~\cite{lin2026factorengine}, graph-structured evolution~\cite{guo2026alphaprobe},
self-evolving agents with experience memory~\cite{factorminer2026, yu2026alphamemo}, safety- and
reproducibility-constrained generation~\cite{shi2026hubble}, market-logic
modeling~\cite{weng2026alphalogics}, and standardized benchmarks~\cite{luo2026alphabench}. Newer
systems add tree search and chain-of-thought prompting~\cite{shi2026alphajungle, cao2025chainofalpha},
multi-agent generation-and-selection pipelines~\cite{chen2025multiagentalpha, vu2026selfimproving}, and
the fusion of formulaic factors with textual newsflow~\cite{guo2025synergy}. The
classic formulaic-alpha vocabulary~\cite{kakushadze2016formulaic} underlies all of these. Our Part~I
sits within this line and shares its proposal-first, memory-driven design. AQuA places this
factor-discovery system alongside a separate autonomous model-development system. The two do not
share agents, memory, or search state; rather, each independently uses prior experimental outcomes to
improve subsequent proposals in its own domain.

\paragraph{Autonomous research agents.}
Beyond finance, language-model agents have been built to run the scientific process end to
end~\cite{lu2024aiscientist, romera2024funsearch, zhou2025autonomoussci, xin2026eurekagent}. The same
agentic pattern has moved into trading, where multi-agent teams of language models propose and act on
investment decisions~\cite{miyazaki2026investteams, singhi2025bitcoinagent}. We adopt the same
ambition of an autonomous research loop, but study recursive self-improvement separately in two
quantitative settings: factor discovery and model development. In each setting, validated evidence
from one iteration is retained and used to improve later research decisions. We additionally target
the specific failure mode of quantitative work, data leakage, by making the data path and the
evaluator unreachable from the agent rather than relying on its judgment.

\paragraph{Self-evolving agents and reusable capabilities.}
Alita~\cite{qiu2025alita} builds, refines, and reuses tools with minimal predefined structure.
AgentDistill~\cite{qiu2025agentdistill} transfers teacher-generated task-solving modules to
student agents without parameter training, while Alita-G~\cite{qiu2025alitag} abstracts successful
trajectories into a reusable library of domain-specific tools. These approaches connect to AQuA's
use of persistent research state to inform subsequent proposals. GenEnv~\cite{guo2025genenv}
studies a complementary feedback loop in which agents and generative environment simulators
co-evolve through difficulty-adapted tasks. AQuA focuses on adapting quantitative research
proposals within fixed data and evaluation contracts.

\paragraph{Live forecasting and financial-agent evaluation.}
FutureX~\cite{zeng2025futurex} evaluates agents on continuously updated future-prediction tasks,
and FutureX-Pro~\cite{liu2026futurexpro} extends this approach to specialized domains including
finance. CryptoBench~\cite{guo2025cryptobench} evaluates expert-designed cryptocurrency retrieval
and prediction tasks, documenting a gap between information gathering and predictive analysis.
These benchmarks emphasize timely evidence and evaluation against subsequently observed outcomes.
AQuA evaluates autonomous research through held-out quantitative signals and trading performance,
with the search process constrained by sealed data paths and separate selection and reporting metrics.

\paragraph{Deep models for financial time series.}
The model in Part~II draws on standard sequence-modeling components, including convolutional
networks for sequences~\cite{bai2018empirical}, state-space models~\cite{gu2023mamba}, and
attention~\cite{vaswani2017attention}, and on deep architectures designed for financial data~\cite{zhang2019deeplob, gu2020empirical}. A recent
wave applies hybrid recurrent, convolutional, and transformer models, often combined with
reinforcement learning, graph, or foundation-model components, to stock-return and portfolio
prediction~\cite{kabir2025lstmtransformer, liu2026transformerclassical, song2025transformerrl, ashrafzadeh2025portfolio, wang2025tgnnllm, marconi2026tsfm, kirtac2025sentiment}. Our contribution is not a new primitive but a separate autonomous loop that lets the
agent compose these primitives into models and accumulate evidence across variants. Its relation to
Part~I is conceptual---both systems learn from prior experiments---rather than architectural.

\section{Method Overview}
\label{sec:method}

Part~I and Part~II of AQuA are separate research systems. They do not share agents, memories,
candidate spaces, or research state. We describe them together here only because both exhibit the
same high-level pattern: each system uses validated evidence from earlier experiments to guide later
research decisions in its own domain.

Within each part, an iteration proceeds through five stages. It begins with a \emph{hypothesis}: a
candidate factor in Part~I, or a model configuration in Part~II. The hypothesis is \emph{constructed
or trained} into a concrete artifact, \emph{evaluated} on held-out data, and \emph{validated}
against look-ahead bias and regime dependence. Surviving artifacts are then \emph{selected or
combined} into that part's running output, a combined factor signal in Part~I and a trading model in
Part~II. A final \emph{persistent research-state update} writes the iteration's evidence to the
store belonging to that part, which its next iteration consults before proposing a new hypothesis.
The next iteration reads the recorded candidates, evaluation results, and proposed follow-up tests.

For part $p \in \{\mathrm{I},\mathrm{II}\}$, let $R_t^{(p)}$ denote its persistent research state
before iteration $t$, $H_t^{(p)}$ its hypothesis, $C_t^{(p)}$ the constructed factor or trained
model, and $E_t^{(p)}$ the validated evidence returned by the experiment. The within-part recursion
is
\begin{equation}
  R_{t+1}^{(p)} = \mathcal{U}_p\!\left(R_t^{(p)}, H_t^{(p)}, C_t^{(p)}, E_t^{(p)}\right),
  \qquad
  \left(H_{t+1}^{(p)}, C_{t+1}^{(p)}\right)
  \sim \mathcal{P}_p\!\left(\cdot \mid R_{t+1}^{(p)}\right).
\end{equation}
The part-specific update $\mathcal{U}_p$ converts experimental outcomes into reusable research
knowledge, and $\mathcal{P}_p$ uses that knowledge to guide the next proposal. The superscript
emphasizes the separation: Part~I does not update $R^{(\mathrm{II})}$, and Part~II does not update
$R^{(\mathrm{I})}$. We use \emph{recursive self-improvement} in this bounded,
research-process-level sense within each part; neither system updates the underlying language model
or the evaluator.

The commonality is therefore a high-level research pattern, not a shared implementation.
Construction writes a symbolic factor expression in Part~I and trains a hybrid time-series model in
Part~II; evaluation reports an information coefficient in both, but under part-specific conventions,
so the two coefficients are not directly comparable and we report them separately.

\subsection{Sealed sandbox and constrained iteration}
\label{sec:sandbox}

Because the language model proposes and, in places, writes its own experiments, the central failure
mode is data leakage: an autonomously generated feature, label, or normalization that consults
information unavailable at prediction time. Backtest overfitting of this kind is well documented and
produces simulated performance that does not survive out of sample~\cite{bailey2014pseudo, bailey2017probability, harvey2016cross}. We restrict the agent to a fixed sandbox before any iteration begins.
The data splits $\mathcal{D}$, the feature and label definitions $\mathcal{F}$ and $\mathcal{L}$,
and the evaluator $\mathcal{V}$ are sealed and human-authored, and the model never edits them. Each
action of the model is a specification $\theta$ drawn from a constrained space $\Theta$, which the
harness compiles and scores through the sealed evaluator,
\begin{equation}
  s_k = \mathcal{V}\big(\mathcal{C}(\theta_k);\, \mathcal{S}\big),
  \qquad \theta_k \in \Theta, \quad
  \mathcal{S} = (\mathcal{D}, \mathcal{F}, \mathcal{L}, \mathcal{V}) \ \text{sealed}.
\end{equation}
The space $\Theta$ is defined so that no $\theta$ can alter $\mathcal{S}$. The restricted interface prevents generated specifications from modifying the data pipeline,
split logic, or evaluator.

Candidate selection creates a separate risk: a search that can read the metric it will ultimately report will, given enough
iterations, learn to select for it. We reserve a final evaluation window separately from the validation slice used to rank candidates. During search the harness returns to the agent only a score
on a validation slice fixed in advance; the score $s_k$ that ranks candidates and the
inner-validation signal that drives early stopping and checkpoint choice are computed on that slice
alone. Where a part designates a final test window, that window is scored once, after the
configuration is frozen, and is never returned to the agent or used to rank candidates. In Part~II
this window is the untouched 2021--2025 period, so the reported test coefficient is out of sample
with respect to the entire search.

In Part~I, the registry $\mathcal{O}$ is the standard vocabulary of formulaic-alpha
operators~\cite{kakushadze2016formulaic, zhang2020autoalpha}. Its leaves are raw fields (open, high,
low, close, volume, vwap, returns); its inner nodes are operators of three kinds: cross-sectional
operators that act across the universe at a fixed timestamp (rank, z-score, sector neutralization),
time-series operators that summarize a trailing window for each entity (lag, difference, moving
correlation and covariance, rolling rank, rolling standard deviation, linear-decay weighting), and
element-wise arithmetic and conditionals. A factor is a composition of these operators over the raw
fields, represented as an expression tree~\cite{zhang2020autoalpha, cui2021alphaevolve}; the running
example
\begin{equation}
  f = \operatorname{rank}\!\big(\operatorname{corr}(r^{1d}, v^{1d}, 20)\big)
    - \operatorname{rank}\!\big(\operatorname{std}(r^{1d}, 20)\big)
\end{equation}
is one such tree. Every time-series operator reads only its trailing window and every
cross-sectional operator reads only the current timestamp, so causality is closed under
composition: any expression the model assembles from $\mathcal{O}$ is causal by construction, and
it cannot introduce a primitive that consults future data. Operator-based search of this space has
a long line of work, from genetic programming~\cite{zhang2020autoalpha, cui2021alphaevolve} and
reinforcement learning~\cite{yu2023synergistic, zhao2025quantfactor} to recent language-model agents~\cite{han2026quantaalpha, tang2025alphaagent, shi2026alphajungle}.
Before assembling the expression, the agent states each factor as a falsifiable proposal, with a
hypothesis, mechanism, predicted direction, and refutation conditions. Listing~\ref{lst:part1-proposal}
shows one such proposal.

\begin{lstlisting}[float=t,
  caption={A proposal in Part~I is a falsifiable factor hypothesis, with its mechanism, predicted
  direction, and refutation conditions, stated before any expression is built. The deployed
  expression is withheld; the full iteration record is in Appendix~\ref{app:part1}.},
  label={lst:part1-proposal}]
proposal:
  hypothesis: >
    After a forced open-interest unwind, a price rebound that is not confirmed
    by aggressive taker flow is more likely to fail.
  mechanism: >
    Deleveraging removes forced pressure, but weak buy-flow and poor basis
    recovery indicate insufficient demand behind the rebound.
  expected_direction: higher_signal_predicts_lower_future_return
  expected_label: [ret_open_open_h10, ret_open_open_h30]
  falsification_criteria:
    - no IC concentration inside the deleveraging event window
    - sign flips or vanishes out of sample or across regimes
    - subsumed by a price, open-interest, or taker-flow baseline
  expected_failure_modes:
    - rebound strength already captured by short-horizon momentum
    - taker-flow gap is noise once the basis has normalized
  factor_blueprint:
    - deleveraging_intensity: ranked negative change in open interest
    - rebound_strength: short-horizon price recovery after the event
    - flow_gap: lack of taker-flow confirmation during the rebound
  expression: withheld
\end{lstlisting}

In Part~II, the specification is a configuration in a domain-specific language whose registry covers
the full experiment: the data split selected from the frozen set, the sampler, the architecture
blocks, the loss, and the optimizer. A hypothesis is a single config diff. The registry is the
model's only surface; it selects and parameterizes registered operators but cannot write the data
loader, the split logic, or the evaluator. Listing~\ref{lst:config} shows one such configuration.
The architecture registry composes standard sequence-modeling primitives, including temporal
convolutions~\cite{bai2018empirical}, state-space mixers~\cite{gu2023mamba}, and
attention~\cite{vaswani2017attention}, into deep multi-resolution stacks.

\begin{lstlisting}[float=t,
  caption={A hypothesis in Part~II is one config diff over the sealed sandbox, the only artifact the
  model emits. Sealed components (splits, features, labels, evaluator) are referenced by id and
  never redefined; operator names and values are illustrative.},
  label={lst:config}]
sandbox:
  splits:    frozen/sp_A          # train / val / test
  features:  frozen/feat_g12
  label:     frozen/lbl_fwd
  evaluator: frozen/eval          # held-out, two-leg cost, walk-forward

input:   {normalize: per_entity_z, stats_window: train_only, history: 64}
sampler: {kind: stratified_minute, batch: 8192}

arch:                             # composed from the architecture registry
  - conv_stem:       {scales: [3, 5, 15]}           # multi-scale 1-D conv
  - multi_resolution:
      fine:   {repeat: 4, block: [temporal_conv, sequence_mixer/state_space]}
      coarse: {repeat: 2, block: [sequence_mixer/attention, feedforward]}
      fuse:   cross_attention
  - block_repeat:
      times: 3                                       # panel interaction
      block: [cross_entity_mixer, temporal_conv/depthwise, feedforward]
  - readout:         {gate: [fine, coarse, panel], pool: [last, mean, attn]}

loss:  [spearman_ic, huber_csz, turnover_reg]
optim: {adamw, lr: 3.0e-4, schedule: cosine, precision: bf16, ddp: 8}
eval:  {cost_bps: 2, walk_forward: expanding}   # held-out
\end{lstlisting}

Formally, $\Theta_{\mathrm{II}} = \{\, c = (\mathrm{split}, \mathrm{sampler}, \mathrm{arch},
\mathrm{loss}, \mathrm{optim}) \,\}$, where $\mathrm{split}$ is chosen from the frozen $\mathcal{D}$
and cannot be redefined, and the remaining components are drawn from their registries. The compiler
builds the model and training run, and the sealed evaluator reports held-out IC, $R^2$, and Sharpe
under a two-leg cost. Every $c \in \Theta_{\mathrm{II}}$ uses the registered features and fixed split logic.
The harness applies the same data-access rules to each configuration. One config diff also corresponds to exactly one variant, which keeps variants directly
comparable. Sections~\ref{sec:part1} and~\ref{sec:part2} instantiate these two registries in full.

\section{Part I: Autonomous Factor Discovery}
\label{sec:part1}

\subsection{System architecture}

\begin{figure}[t]
  \centering
  \includegraphics[width=\textwidth]{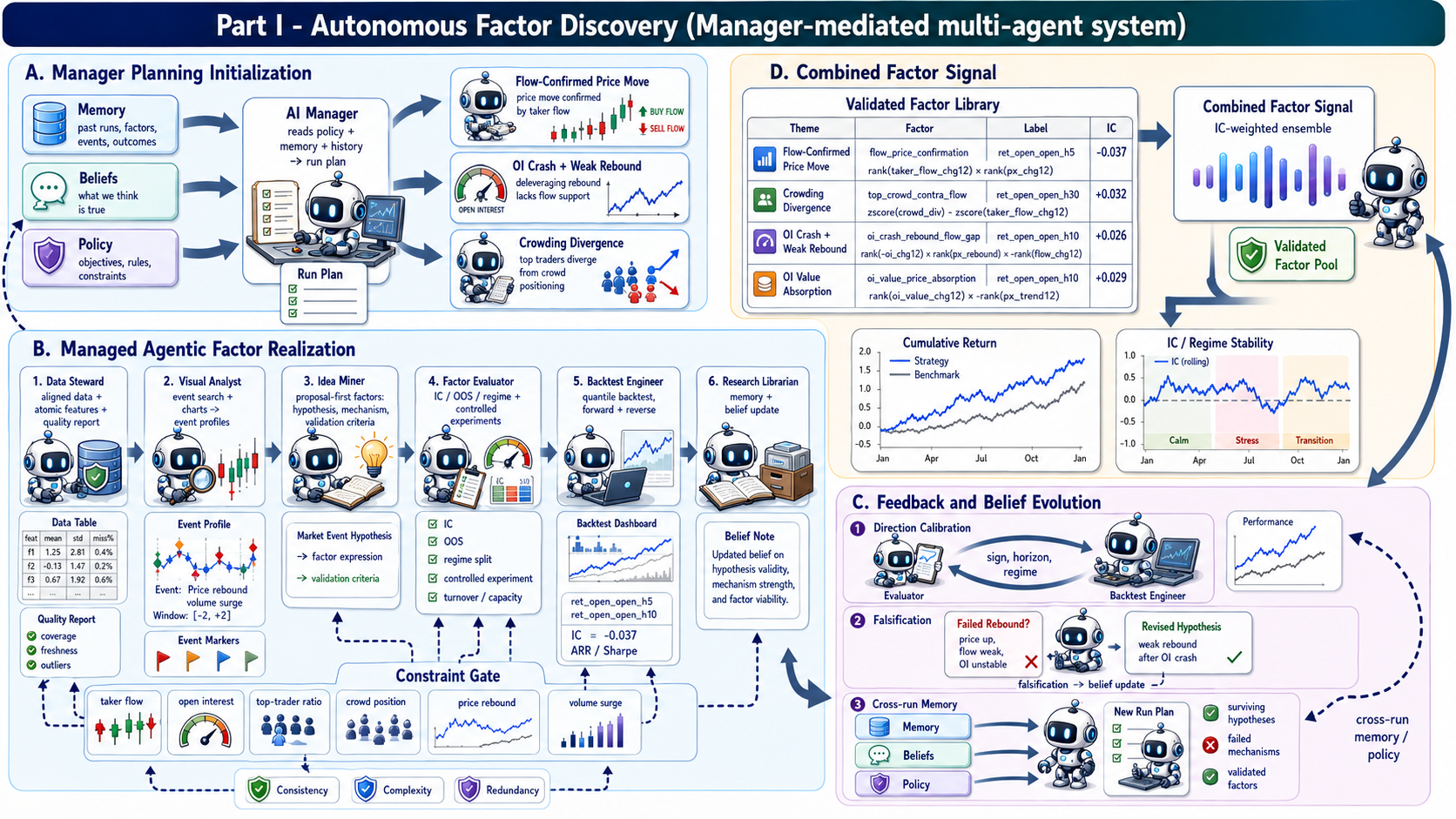}
  \caption{Part~I architecture. An AI Manager orchestrates a six-agent pipeline (Data Steward
  $\to$ Visual Analyst $\to$ Idea Miner $\to$ Factor Evaluator $\to$ Backtest Engineer $\to$
  Research Librarian) that turns a research goal into a combined factor signal. Three nested
  feedback loops, for direction calibration, falsification-driven belief update, and cross-run
  memory and policy, drive iterative improvement, backed by persistent Memory, Beliefs, and
  Policy stores.}
  \label{fig:part1-arch}
\end{figure}

Part~I of AQuA is a multi-agent system that turns a research goal into validated factors. An AI Manager
mediates each run. It reads the research policy, the accumulated memory, and the record of previous
runs, and from these it writes a plan that assigns every downstream agent a concrete task. The
agents never call one another; each handoff passes through the Manager, which keeps a run auditable
and reproducible. Figure~\ref{fig:part1-arch} shows the pipeline.

Six specialist agents run in sequence. The Data Steward loads and aligns the market data and
returns a quality report together with a set of atomic features. The Visual Analyst searches the
history for representative events of a requested type and summarizes them into event profiles. The
Idea Miner proposes candidate factors from those profiles. The Factor Evaluator scores each
candidate, the Backtest Engineer trades it in simulation, and the Research Librarian records the
outcome. The proposal and validation steps are detailed below; the specific factors the system
selects are not disclosed.

\subsection{The research pipeline}

A factor enters the pipeline as a proposal rather than as an expression. For each candidate the
Idea Miner states a hypothesis, the economic mechanism behind it, the direction it is expected to
predict, and the conditions under which it should be considered refuted. This framing requires
every factor to carry its own rationale and its own test before it is built, a discipline that the
recent agentic-mining literature has converged on~\cite{han2026quantaalpha, alphaagentevo2026,
shi2026hubble, tang2025alphaagent, vu2026selfimproving}. The candidate is then assembled from the
formulaic-alpha operator registry of Section~\ref{sec:sandbox}, where Listing~\ref{lst:part1-proposal}
shows the proposal form it emits.

Evaluation follows the same contract for every proposal. The Factor Evaluator computes information
coefficients across forward-return labels, monthly stability, held-out split behavior, market-regime
behavior, turnover, expression complexity, and correlation with the existing factor pool. It also
performs controlled comparisons against simple baselines built from price, volume, open interest,
basis, and taker-flow changes. When a proposal is tied to an event, its effect inside the event
window is compared with a control region outside the event. A factor is carried forward when its
signal is not simply a restatement of a baseline and when its strongest performance appears in the
market context predicted by its mechanism.

The Backtest Engineer then turns the evaluated signal into a simple quantile portfolio. It tests
both the proposed direction and the reversed direction and keeps the cleaner trading interpretation.
This direction-calibration step matters because a formula can have predictive content even when the
initially stated sign is wrong. The final object stored by the system is therefore not only a factor
score, but a factor together with its direction, target horizon, supporting mechanism, evaluation
evidence, and trading behavior.

\subsection{Cross-run learning}

After each run, the Research Librarian
writes a structured record containing the goal, event type, visual observations, proposed
mechanisms, evaluated factors, selected signals, backtest summaries, and updated beliefs. A belief
attaches confidence to a mechanism in a market context, such as ``open-interest crashes followed by
weak flow-confirmed rebounds tend to reverse'' or ``quiet volume bursts continue only when price
acceptance and taker flow agree.'' The next run is planned against this memory.

This creates three feedback loops. The first operates within a single backtest, where the system
calibrates the direction of a signal. The second operates within a run, where failed proposals
update the recorded confidence in the proposed mechanism. The third operates across runs,
where the AI Manager reads the accumulated memory and steers the next search toward mechanisms supported by earlier evaluations. In practice, later runs do not start from a blank prompt. They inherit prior
observations, avoid mechanisms that repeatedly failed, and refine promising mechanisms with new
event definitions, horizons, or conditioning variables.

\subsection{A worked iteration}

To make the loop concrete, Appendix~\ref{app:part1} records one full iteration, and we summarize it
here. The run asks whether a weak price rebound after an open-interest crash predicts reversal. The
Manager turns this question into a plan: find deleveraging episodes, inspect price and flow after the
unwind, and test whether rebound quality separates continuation from failure. The Visual Analyst
returns event profiles that separate a clean forced-deleveraging pattern, where open interest falls,
volume expands, and price rebounds only briefly, from a healthier reset where taker flow and basis
recover with price. From this the Idea Miner proposes a mechanism in which a rebound left unconfirmed
by aggressive taker flow is more likely to fail. The evaluator clears it against the price,
open-interest, and flow baselines, with its skill concentrated in the event window its mechanism
targets. The strongest factors in this family reach single-factor information coefficients on the
order of $0.026$ to $0.037$. A later run changes the goal to quiet-market volume expansion and reuses
the same machinery without touching the evaluator, showing that the loop explores a new mechanism by
re-planning rather than re-coding. The deployed expressions are withheld throughout.

\subsection{Results}

\begin{figure}[t]
  \centering
  \includegraphics[width=0.9\textwidth]{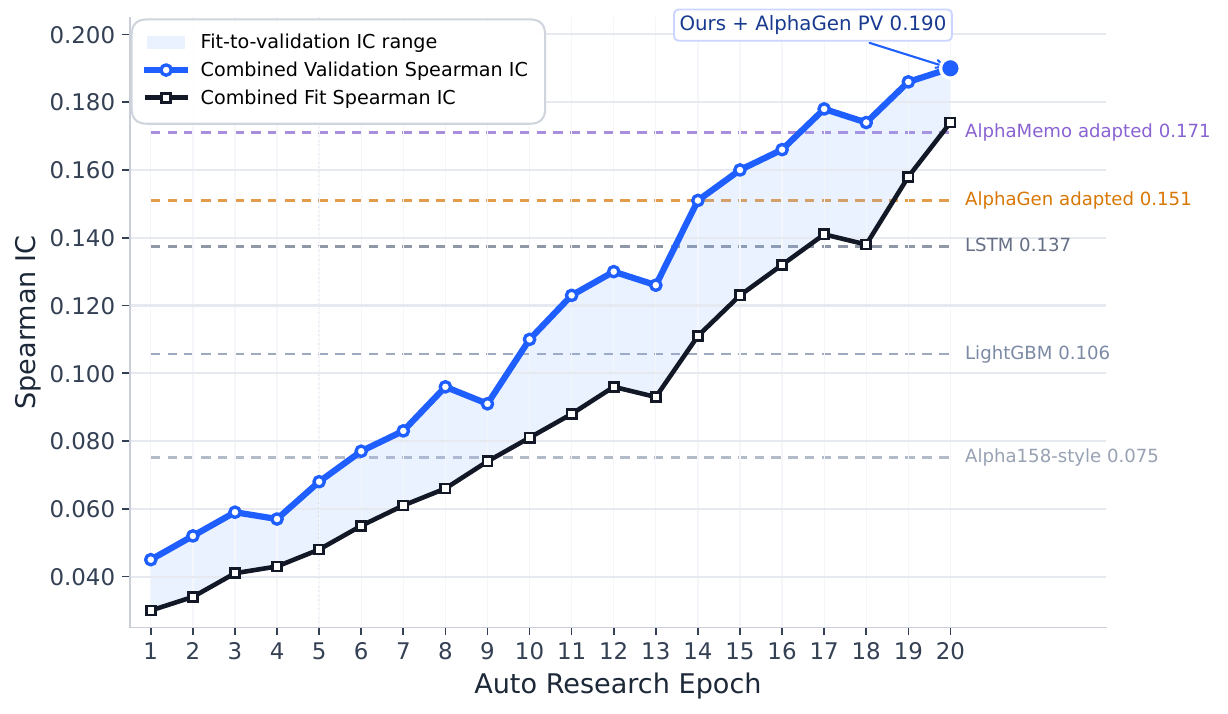}
  \caption{Part~I combined-factor signal across autonomous-research iterations. The combined
  validation IC improves as the loop accumulates and reuses validated evidence, reaching a combined
  signal IC of approximately $0.190$. \emph{Part~I IC is the combined-factor Spearman IC on the
  crypto five-minute universe, a different convention from Part~II; the two numbers should not be
  compared directly.}}
  \label{fig:part1-progression}
\end{figure}

The output of Part~I is a combined factor signal formed from the factors that survive the evaluation
and direction-calibration stages. Figure~\ref{fig:part1-progression} reports the quality of this
combined signal across autonomous-research iterations. The combined validation information
coefficient rises as the loop accumulates and reuses evidence, reaching approximately $0.190$. The
improvement coincides with the addition of richer event profiles, open-interest and flow
conditioning, crowding-divergence mechanisms, and regime-aware factor selection. As noted in
Section~\ref{sec:method}, the Part~I information coefficient follows the combined-factor convention
on the crypto five-minute universe, and is not comparable to the per-stock coefficient reported in
Part~II.

Selected factors encode proposed economic mechanisms and typically have an
information coefficient around $0.03$ in absolute value. This is the expected regime for intraday
formulaic signals: the strength comes not from any single rule but from combining a library of
mechanism-grounded factors into a stronger aggregate signal. The Part~I result is
therefore not a claim that one discovered expression is sufficient, but that an autonomous,
memory-bearing research harness can repeatedly turn market hypotheses into tested factor evidence and
improve the combined signal over iterations.

\section{Part II: Autonomous Model Development}
\label{sec:part2}

\subsection{The research loop}

\begin{figure}[t]
  \centering
  \includegraphics[width=\textwidth]{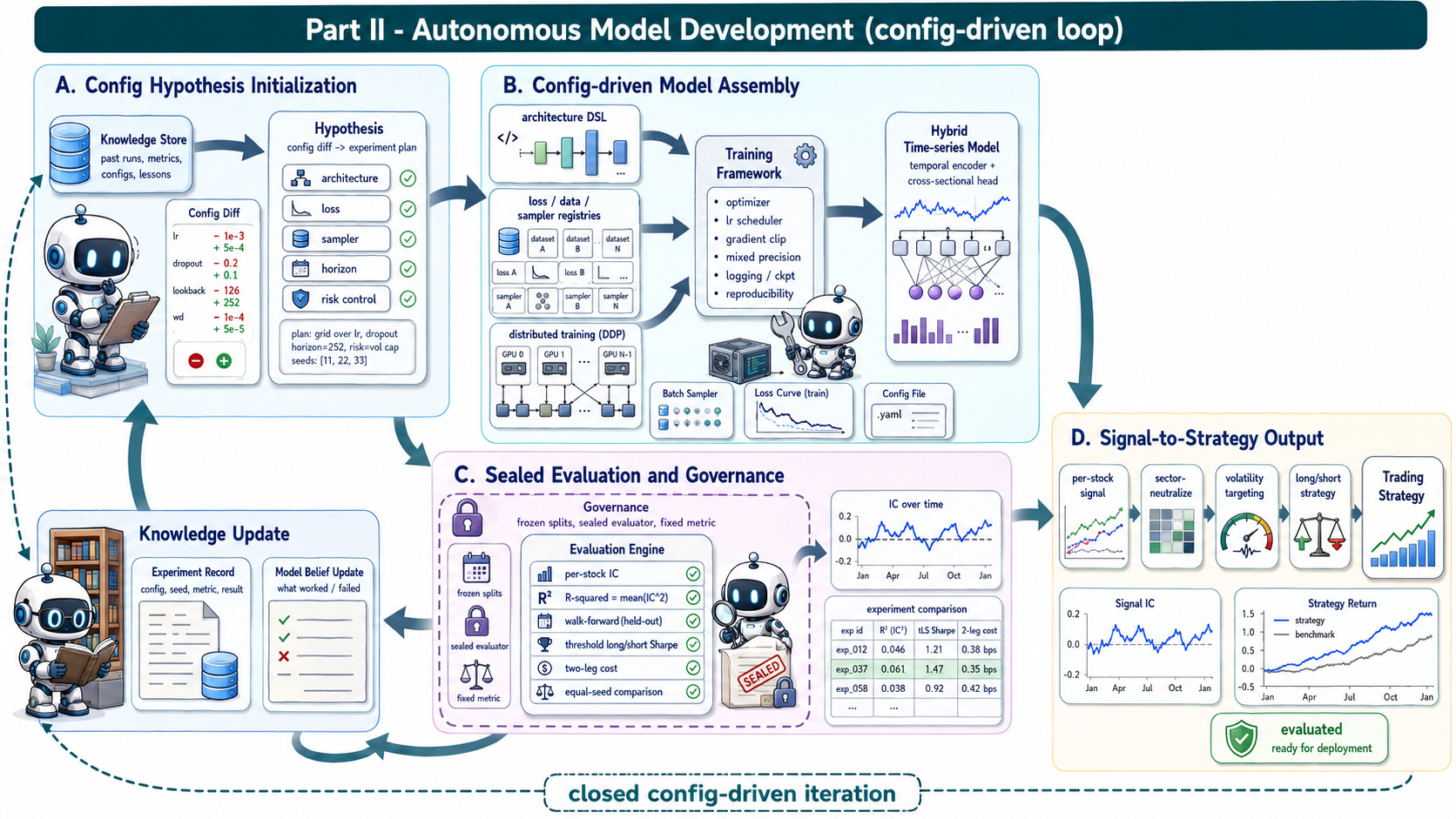}
  \caption{Part~II architecture. A config-driven loop, from a hypothesis (config diff) through the
  training framework and the evaluation engine to a knowledge update, iterates over model variants.
  The training framework (architecture DSL, loss, data, and sampler registries, distributed
  training) produces a hybrid time-series model; the evaluation engine reports held-out per-stock
  IC, $R^2=\mathrm{mean}(IC^2)$, walk-forward, and a two-leg-cost threshold long/short Sharpe,
  within a governance contract of frozen splits, a sealed evaluator, and a
  fixed selection metric. The model signal is constructed into a trading strategy
  (sector-neutralize, then volatility targeting, then long/short).}
  \label{fig:part2-arch}
\end{figure}

Part~II of AQuA develops trading models through the config-driven loop of Figure~\ref{fig:part2-arch}. Each
iteration begins with a hypothesis written as a single config diff over the sealed sandbox of
Section~\ref{sec:sandbox}: a change to the architecture, the loss, the sampler, or the optimizer.
The harness compiles the diff into a training run, scores it through the sealed evaluator, and
writes the result to a knowledge store that the next hypothesis reads. Because one diff produces
exactly one variant and the data path is fixed, two variants differ only in the knobs that changed.
The same data and evaluation protocol are applied to every variant.

\subsection{The hybrid model}

\begin{figure}[t]
  \centering
  \includegraphics[width=\textwidth]{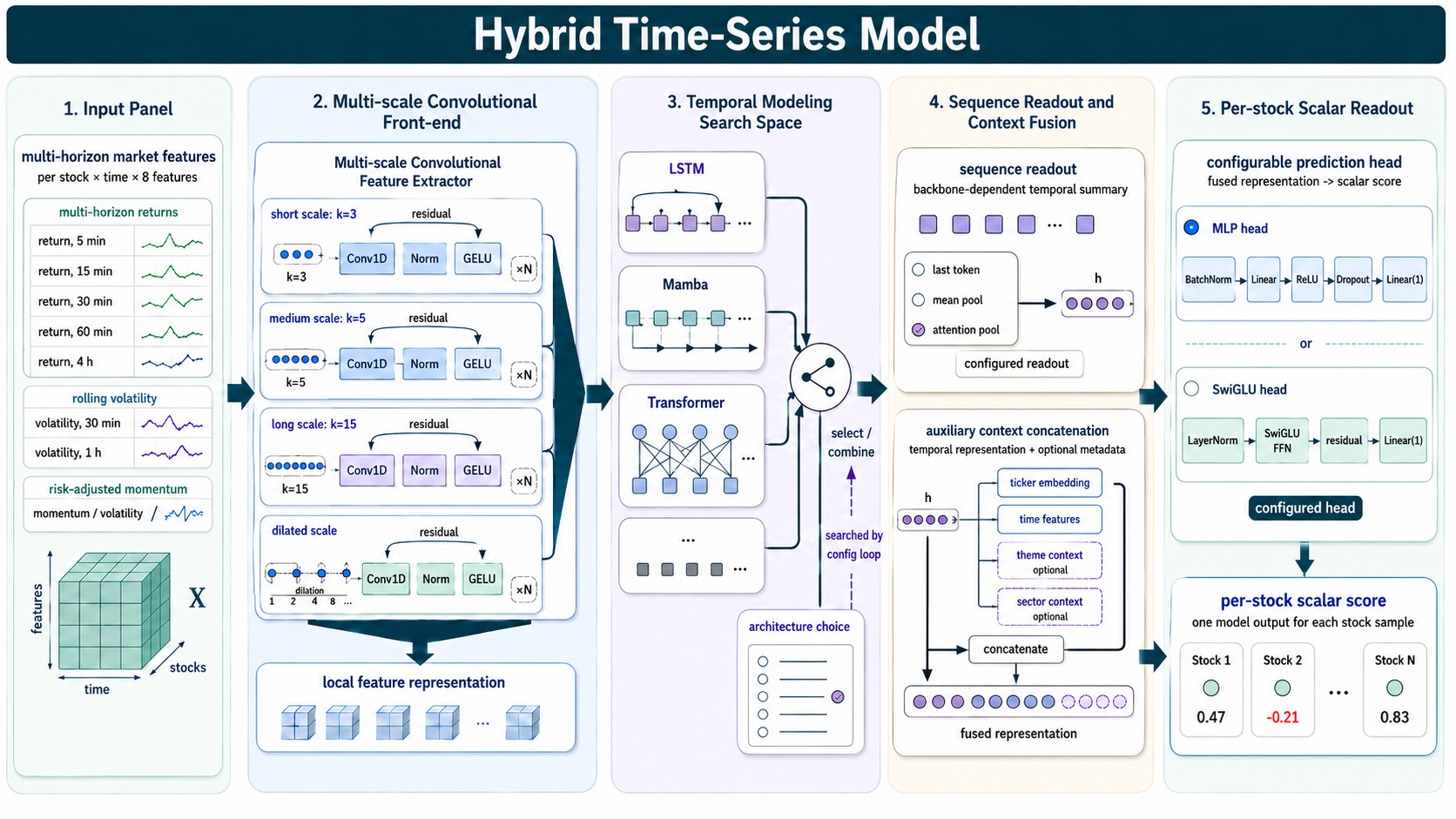}
  \caption{
The hybrid time-series model. A multi-scale convolutional front-end extracts
local representations from multi-horizon return, volatility, and
risk-adjusted momentum features at each timestep. A configurable temporal
backbone, instantiated as a recurrent, state-space, or attention-based
sequence model, captures longer-range temporal dynamics. A backbone-dependent
sequence readout produces a fixed-dimensional representation, which is
concatenated with configured auxiliary information such as ticker embeddings,
time features, and optional theme or sector context. A configurable MLP or
SwiGLU prediction head then emits one scalar score for each stock sample.
}
  \label{fig:hybrid}
\end{figure}

Our predictor is a hybrid model that combines convolutional feature extraction with sequence
modeling~\cite{zhang2019deeplob, kabir2025lstmtransformer}, shown in Figure~\ref{fig:hybrid}. The front-end is a deep convolutional stack: several
blocks of one-dimensional convolutions run over the input history at multiple kernel sizes and
dilations, building a rich local representation of each stock's recent price-volume dynamics at
every timestep. We leave the precise convolutional configuration unspecified. These representations
feed a temporal-modeling stage instantiated from sequence models, spanning recurrent networks (LSTM~\cite{hochreiter1997lstm}), state-space models (Mamba~\cite{gu2023mamba}), and attention
(Transformer~\cite{vaswani2017attention, bai2018empirical}); the configuration used in our
experiments uses attention. A cross-sectional stage then mixes information across the panel of
stocks at each timestep, the branches are fused with a gating mechanism, and a pooled readout emits
the per-stock score.

The whole model is one point in the configuration space of Section~\ref{sec:sandbox}: the
convolutional front-end, the sequence family, the cross-sectional mixing, and the fusion are all
operators in the registry, so a new model is a new config rather than new code. The framework
trains it under the sealed sandbox, and the exact feature set, normalization, and label
construction are part of that sandbox and are not disclosed.

\subsection{Task, features, and baselines}
\label{sec:part2-baselines}

We instantiate Part~II on intraday US equity prediction. The task is to predict each stock's
forward return over the next thirty minutes. We split the data chronologically: we train on
2010--2019, leave 2020 as an embargo gap that no part of training or selection touches, and report
on the untouched 2021--2025 test window. Model selection (early stopping and checkpoint choice) is
driven only by an inner-validation slice taken from the end of the training window, and the test
window is used solely for final evaluation, never informing training or selection. The model input
is a short history of pure price-volume features, standardized per stock.

No single feature carries the signal on its own. Table~\ref{tab:feat} reports the single-feature
information coefficient of a representative set of price-volume features over the held-out window.
None of these features, nor a ridge linear combination of them, exceeds about $0.03$ in magnitude; the ridge reaches $+0.025$. The hybrid model combines these inputs over time and achieves a higher IC than the tested
single-feature predictors and ridge baseline.

\begin{table}[t]
  \centering
  \caption{Single-feature information coefficient of representative price-volume features, held out
  over 2021--2025 (per-stock raw IC). No single feature, or ridge combination of them, exceeds
  about $0.03$ in magnitude.}
  \label{tab:feat}
  \begin{tabular}{lc}
    \toprule
    Price-volume feature & single-feature IC \\
    \midrule
    return, 5\,min       & $-0.031$ \\
    return, 15\,min      & $-0.021$ \\
    return, 30\,min      & $-0.013$ \\
    return, 60\,min      & $-0.001$ \\
    return, 4\,h         & $+0.002$ \\
    volatility, 30\,min  & $+0.006$ \\
    volatility, 1\,h     & $+0.004$ \\
    momentum / volatility & $+0.007$ \\
    \bottomrule
  \end{tabular}
\end{table}

We compare a range of model families on this task using the same held-out evaluator.
The search and checkpoint-selection scores come from the inner-validation slice described above. Table~\ref{tab:models} reports their per-stock raw IC, from a linear model
through gradient boosting and recurrent
networks~\cite{hochreiter1997lstm, cho2014gru, beck2024xlstm} to our hybrid model. The progression is clear:
linear and tree models capture part of the signal, sequence models more, and our hybrid model is
the strongest.

\begin{table}[t]
  \centering
  \caption{Model comparison on the held-out window (2021--2025), per-stock raw IC. Models are
  trained on identical data and scored by the same evaluator. Our hybrid model is the strongest.}
  \label{tab:models}
  \begin{tabular}{llc}
    \toprule
    Model & Family & raw IC \\
    \midrule
    Linear (ridge)        & linear            & $+0.0251$ \\
    LGB                   & gradient boosting & $+0.0397$ \\
    xLSTM                 & recurrent         & $+0.0434$ \\
    LSTM                  & recurrent         & $+0.0535$ \\
    GRU                   & recurrent         & $+0.0613$ \\
    \textbf{Ours (hybrid)} & \textbf{hybrid}  & $\mathbf{+0.0843}$ \\
    \bottomrule
  \end{tabular}
\end{table}

\subsection{Evaluation engine}

Every run is scored by the same sealed evaluator. It reports three quantities on held-out data: a
per-stock time-series information coefficient, a per-stock $R^2$ defined as the cross-sectional mean
of squared IC, and a threshold long/short Sharpe ratio under a two-leg turnover cost.

\begin{figure}[t]
  \centering
  \includegraphics[width=0.85\textwidth]{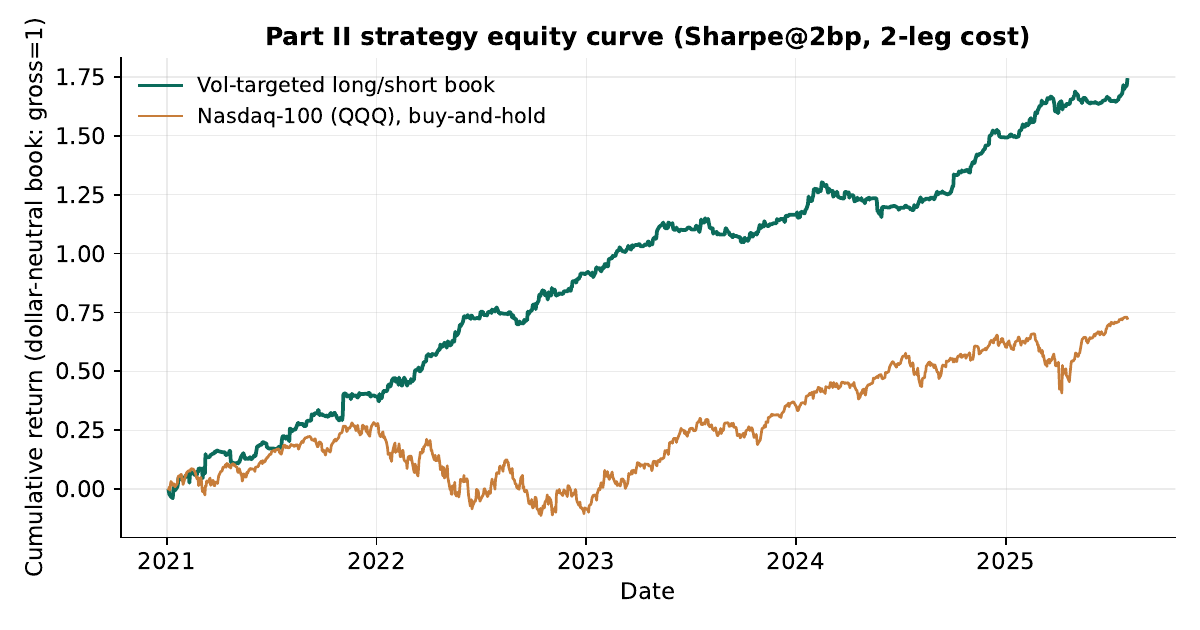}
  \caption{Part~II strategy equity curve. Cumulative return of the volatility-targeted,
  dollar-neutral threshold long/short book (gross exposure $1$, two-leg turnover cost
  $2$\,bps) over 2021--2025, with the Nasdaq-100 (QQQ) buy-and-hold return over the same
  window shown for reference. The market-neutral book compounds more smoothly and sidesteps
  the 2022 index drawdown; QQQ is long-only and not risk-matched to the neutral book.}
  \label{fig:part2-equity}
\end{figure}

Figure~\ref{fig:part2-equity} summarizes the trading result. The cumulative return of the
volatility-targeted, dollar-neutral long/short book climbs steadily across 2021--2025, making new
highs into the end of the window and ending well above a Nasdaq-100 buy-and-hold benchmark over the
same period without taking its 2022 drawdown.

\subsection{From signal to strategy}

We trace one model from prediction to deployable strategy. A trained model produces a per-stock
score at each timestamp. The score is turned into a dollar-neutral threshold long/short book: stocks
above an upper threshold are held long and stocks below a lower threshold short, the book is
rebalanced on a fixed cadence, and each leg pays a two-leg turnover cost. Two construction steps lift
the held-out Sharpe. Sector-neutralizing the score removes common sector exposure and raises the
held-out Sharpe to $+2.15$, with similar values in the training and held-out periods. A causal volatility-targeting overlay, which scales daily exposure
by an online estimate of trailing volatility toward an expanding-median target, raises it further to
$+2.50$.

\subsection{Results}

Table~\ref{tab:part2-main} reports the held-out metrics for 2021--2025. The model reaches a
per-stock raw information coefficient of $+0.0843$ and a per-stock $R^2$ of $1.20\%$. The threshold
long/short book reaches a Sharpe of $+2.15$ before the volatility overlay and $+2.50$ after it, and
a fully causal walk-forward that chooses every parameter from past data alone still reaches $+2.0$.
Table~\ref{tab:part2-byyear} breaks the Sharpe down by year. It is positive in every year from 2021
to 2025, including the 2022 market decline. As stated in Section~\ref{sec:method}, the Part~II information coefficient is a
per-stock time-series quantity, reported on its own and not against Part~I.

\begin{table}[t]
  \centering
  \caption{Part~II main results on the held-out evaluation window (2021--2025). All figures
  are out of sample, with model selection on validation only. The information coefficient is
  the per-stock time-series Pearson IC (raw); $R^2$ is the cross-sectional mean of squared IC;
  the Sharpe ratio is for a dollar-neutral threshold long/short book at a two-leg turnover cost
  of $2$\,bps, with parameters tuned on a training fraction and scored on the held-out
  remainder. \emph{(IC convention differs from Part~I and the two should not be compared.)}}
  \label{tab:part2-main}
  \begin{tabular}{lc}
    \toprule
    Metric (held out, 2021--2025) & Value \\
    \midrule
    \ding{172}~Per-stock IC (raw)                 & $+0.0843$ \\
    \ding{173}~Per-stock $R^2 = \mathrm{mean}(IC^2)$ (raw) & $1.20\%$ \\
    \ding{174}~Sharpe@2bp, sector-neutral book    & $+2.15$ \\
    \quad\, $+$ causal volatility targeting       & $+2.50$ \\
    \quad\, fully-causal walk-forward (no hindsight) & $+2.00$ \\
    \bottomrule
  \end{tabular}
\end{table}

\begin{table}[t]
  \centering
  \caption{Part~II strategy Sharpe@2bp by calendar year, held out over 2021--2025.}
  \label{tab:part2-byyear}
  \begin{tabular}{lccccc}
    \toprule
    Year & 2021 & 2022 & 2023 & 2024 & 2025 \\
    \midrule
    Sharpe@2bp & $+1.7$ & $+3.5$ & $+1.9$ & $+1.8$ & $+2.7$ \\
    \bottomrule
  \end{tabular}
\end{table}

\section{Discussion}
\label{sec:discussion}

AQuA applies iterative experiment design to two research tasks. In factor discovery, the Manager
uses prior evaluations to revise economic hypotheses and factor expressions. In model development,
the agent proposes configuration changes using records of earlier training runs. These records make
failed candidates and promising follow-up experiments available to subsequent iterations.

The restricted interface fixes which parts of an experiment the agent can change. Factor proposals
compose registered operators, and model proposals select registered architecture and training
options. The harness owns the data pipeline, chronological splits, and evaluator. This design
prevents a proposal from replacing the split logic or introducing an arbitrary preprocessing routine,
including the full-day volume normalizer that caused the failure in Appendix~\ref{app:part2-failures}.

Evaluation separates candidate selection from final reporting. Validation scores guide the search;
a reserved test window measures the selected model after its configuration is frozen. Part~I reports
combined-factor validation IC across search iterations. Part~II reports final predictive performance
on 2021--2025 and strategy results under the stated turnover cost, including a walk-forward Sharpe
of about $2.0$. These evaluations address different stages of the research process.

A next experiment is to use the discovered factors as inputs to the model-development loop.
The factor library and its selection dates would become part of the model's data specification.
Freezing that library before model selection, with chronological separation of both searches from
final testing, would preserve a traceable record of which information was available at each stage.

\section{Limitations}
\label{sec:limitations}

Two scope limitations bound these results. Each system is demonstrated on a single market and
horizon, crypto at five minutes for Part~I and US equities at thirty minutes for Part~II, and we do
not claim the numbers transfer to other markets or frequencies without re-tuning. The loops also run
with a human operator who sets the research goal, owns the sandbox, and supervises promotion, so the
systems are autonomous within those bounds rather than unattended. The reported metrics are
simulated under a turnover-cost model and have not been validated in live trading.

The guarantees are also uneven across the two channels of leakage. The restricted interface prevents the agent from changing the data and feature
pipeline. Its temporal correctness still depends on the registered operators, source-data timestamps,
and preprocessing implementation. Keeping the final test window out of selection is weaker. The harness returns only
validation scores during search, but the isolation of the test window rests on the sealed protocol
and operator discipline rather than a hard technical barrier, and an operator with direct access to
the store could in principle consult it. We therefore treat test isolation as a governance property
to be audited over a run, not as a cryptographic guarantee.

The progression across iterations combines memory reuse, additional search, and changes to the
candidate pool. Matched-budget comparisons with memory disabled would measure the contribution
of cross-run memory. Part~I reports validation performance used during search; a separately reserved
test period would measure the selected factor pool after the full search.

\section{Conclusion}
\label{sec:conclusion}

We presented AQuA, which comprises two separate autonomous research systems: one for factor discovery
and one for model development. Each system records evaluated candidates and uses those records to propose subsequent experiments.
A restricted DSL allows changes to factor expressions or model configurations while the harness
maintains the data pipeline and evaluator. On a crypto universe the factor system reaches a
combined validation IC of about $0.190$, and on US equities the
model system reaches a per-stock IC of $+0.0843$ and a threshold long/short Sharpe of
$+2.50$ on the held-out evaluation, with about $+2.0$ under walk-forward evaluation. Coupling the two systems, so that discovered factors feed the model
loop, is the natural next direction.

\bibliography{references,rsi_references,alpha_references}

\appendix
\section{Part~I iteration record and walkthrough}
\label{app:part1}

This appendix gives the full version of the worked iteration summarized in
Section~\ref{sec:part1}. Listing~\ref{lst:part1-iteration} shows the record the system stores for a
single iteration, and the text below traces two iterations in detail.

\begin{lstlisting}[
  caption={One Part~I iteration record. The Idea Miner emits a falsifiable proposal rather than a
  bare formula. The evaluator then attaches controlled tests, direction calibration, and a memory
  update. Variable names are representative; deployed expressions are withheld.},
  label={lst:part1-iteration}]
run:
  goal: "After an open-interest crash, does a weak rebound predict reversal?"
  event_type: open_interest_crash
  universe: BTCUSDT_5m

manager_plan:
  visual_search:
    trigger:
      - sharp_drop(open_interest)
      - sharp_drop(open_interest_value)
      - elevated(volume)
    context_fields:
      - price
      - volume
      - basis_rate
      - taker_buy_sell_ratio
      - long_short_account_ratio
      - top_trader_position_ratio
  target_labels: [ret_open_open_h10, ret_open_open_h30]
  instruction: "Test whether rebound quality after deleveraging separates continuation from failure."

visual_observation:
  event_profile:
    - open interest falls abruptly during a high-volume unwind
    - price often rebounds after forced pressure fades
    - rebounds with weak taker-flow confirmation frequently stall
    - basis recovery and positioning reset separate clean rebounds from failed ones

proposal:
  proposal_id: oi_crash_weak_rebound_flow_gap
  hypothesis: >
    After a forced open-interest unwind, a price rebound that is not confirmed
    by aggressive taker flow is more likely to fail.
  mechanism: >
    Deleveraging removes forced pressure, but weak buy-flow and poor basis
    recovery indicate insufficient demand after the rebound.
  expected_direction: higher_signal_predicts_lower_future_return
  factor_blueprint:
    - deleveraging_intensity: ranked negative change in open interest
    - rebound_strength: short-horizon price recovery after the event
    - flow_gap: lack of taker-flow confirmation during the rebound
    - basis_filter: weak or compressed basis-rate recovery
  expression: withheld

evaluation_contract:
  primary_labels:
    - ret_open_open_h10
    - ret_open_open_h30
  baselines:
    - short_horizon_price_momentum
    - open_interest_change
    - taker_flow_change
    - basis_rate_change
  controlled_tests:
    - event_window_ic_vs_control_window_ic
    - full_factor_vs_best_component
    - monthly_ic_stability
    - correlation_with_existing_factor_pool
    - proposed_direction_vs_reversed_direction

structured_observation:
  verdict: selected_for_factor_pool
  single_factor_ic_range: approximately_0.026_to_0.037
  finding: >
    The mechanism is strongest when the open-interest shock is followed by
    weak rebound acceptance and poor taker-flow confirmation.
  memory_update:
    belief: "OI crash rebounds without flow confirmation are more likely to fail."
    action: "Increase priority of deleveraging-plus-flow-gap mechanisms in later runs."
\end{lstlisting}

Listing~\ref{lst:part1-iteration} traces one iteration from a research question to a stored belief.
The run asks whether a weak rebound after an open-interest crash predicts reversal. The AI Manager
converts this question into a concrete plan: search for deleveraging episodes, inspect price and
flow behavior after the unwind, and test whether rebound quality separates continuation from
failure.

The Visual Analyst returns event profiles rather than a single chart. Some episodes show a clean
forced-deleveraging pattern: open interest falls quickly, volume expands, basis compresses, and price
rebounds only briefly. Other episodes show a healthier reset, where taker flow and basis recover
with price. The Idea Miner uses this distinction to propose mechanisms such as weak rebound after
deleveraging, flow-confirmed continuation, and crowded-position unwind. The exact expressions are
withheld, but the generated factors combine open-interest shocks, short-horizon price response,
taker-flow imbalance, basis behavior, and positioning divergence.

The Factor Evaluator then scores each proposal against multiple horizons. In this family of runs,
the strongest open-interest-crash example selected by the system was an ``OI crash rebound flow
gap'' mechanism, which tests whether a rebound after an open-interest shock is unsupported by
aggressive flow. Related proposals tested whether open-interest value rises without price
acceptance, whether top-trader positioning diverges from broader account ratios, and whether
taker-flow confirmation changes the sign of short-horizon price continuation. The best
single-factor examples in this family reached information coefficients on the order of
$0.026$--$0.037$ depending on the label and event context, and the selected signals were then passed
to the combination layer.

A second iteration illustrates how the same loop changes research direction. The goal is changed to
quiet-market volume expansion: the system searches for low-volatility, low-activity periods followed
by a sudden burst in volume, quote volume, and trade count. The Visual Analyst rejects generic
already-volatile high-volume cascades and focuses on true quiet-to-active transitions. The Idea
Miner then proposes continuation factors when the burst is accepted by price, taker flow, and open
interest, and reversal factors when the burst has a weak candle body, noisy flow, or no basis
confirmation. This shows how the same architecture can explore a new market mechanism without
changing the underlying evaluator.

\section{A Part~II failure case: leakage that survived agent review}
\label{app:part2-failures}

The sealed sandbox of Section~\ref{sec:sandbox} is not the design AQuA started from. It is the
response to concrete failures of an earlier, more permissive loop in which the agent could write
feature and factor code directly and a second agent reviewed each candidate for look-ahead bias. We
record the most instructive failure here, because it is the reason AQuA constrains the agent to a
fixed operator registry rather than trusting review.

In the earlier loop, the agent authored each feature as code and a separate reviewer agent checked it
for causality before training. One proposed feature was an intraday volume-participation ratio: the
volume traded from the open up to the current minute, divided by a daily volume normalizer. The
intent is causal and the description reads as backward-looking, so the reviewer agent approved it.
The implementation, however, normalized by the current day's total volume, a sum that runs from the
open through the close. The denominator therefore depended on bars after the current minute, and the
feature quietly encoded end-of-day information into every intraday timestamp. The same failure
appeared in a multi-resolution variant whose daily branch aggregated all of the current day's bars
and was then read at mid-day timestamps.

The symptom was a held-out information coefficient far above what comparable price-volume features
produced, and it did not survive a clean re-split of the evaluation window. A manual audit traced it
to the full-day denominator. The reviewer agent had reasoned about the feature's economic intent, a
ratio of past volume, rather than the exact set of bars its implementation touched, a blind spot it
shared with the author agent.

The lesson is that an LLM reviewing LLM-written code is advisory, not structural: the author and the
reviewer share the same failure modes, so a subtle temporal-footprint bug can pass both. AQuA's
response is operatorization. The agent no longer writes feature or factor code. It composes a fixed
registry of causal operators in which every time-series operator reads only a trailing window ending
at the current timestamp and every cross-sectional operator reads only the current timestamp.
Causality is then closed under composition, as described in Section~\ref{sec:sandbox}, and a full-day
normalizer is not expressible in the specification space at all. The restricted registry excludes this full-day normalization operation from agent-generated specifications.

\end{document}